\documentclass[10pt,twocolumn,letterpaper]{article}

\usepackage[pagenumbers]{cvpr} %

\usepackage{graphicx}
\usepackage{amsmath}
\usepackage{amssymb}
\usepackage{booktabs}
\usepackage[group-separator={,}]{siunitx}
\usepackage{multirow}
\usepackage{placeins}
\usepackage{enumitem}
\usepackage{paralist}

\usepackage[table]{xcolor}
\usepackage{pifont}%
\usepackage{overpic}
\usepackage{array}
\usepackage{xfrac}
\newcolumntype{P}[1]{>{\centering\arraybackslash}p{#1}}
\usepackage[pagebackref,breaklinks,colorlinks]{hyperref}

\definecolor{mydarkblue}{rgb}{0,0.08,0.55}
\hypersetup{  %
    colorlinks=true,
    linkcolor=mydarkblue,
    citecolor=mydarkblue,
    filecolor=mydarkblue,
    urlcolor=mydarkblue
}

\usepackage[capitalize]{cleveref}
\crefname{section}{Sec.}{Secs.}
\Crefname{section}{Section}{Sections}
\Crefname{table}{Table}{Tables}
\crefname{table}{Tab.}{Tabs.}

\def\cvprPaperID{1770} %
\def\confName{CVPR}
\def\confYear{2022}

\newcommand{\bc}{\mathbf{c}}
\newcommand{\bd}{\mathbf{d}}

\newcommand{\bo}{\mathbf{o}}
\newcommand{\bp}{\mathbf{p}}\newcommand{\bP}{\mathbf{P}}

\newcommand{\br}{\mathbf{r}}\newcommand{\bR}{\mathbf{R}}

\newcommand{\bt}{\mathbf{t}}

\newcommand{\bx}{\mathbf{x}}

\newcommand{\nR}{\mathbb{R}}
\newcommand{\nS}{\mathbb{S}}

\newcommand{\cL}{\mathcal{L}}

\newcommand{\cN}{\mathcal{N}}

\newcommand{\cR}{\mathcal{R}}
\newcommand{\cS}{\mathcal{S}}

\newcommand{\figref}[1]{Fig.~\ref{#1}}
\newcommand{\secref}[1]{Sec.~\ref{#1}}

\newcommand{\tabref}[1]{Tab.~\ref{#1}}

\makeatletter
\DeclareRobustCommand\onedot{\futurelet\@let@token\@onedot}
\def\@onedot{\ifx\@let@token.\else.\null\fi\xspace}
\def\eg{e.g\onedot} 
\def\ie{i.e\onedot}

\def\etal{et~al\onedot}

\makeatother

\definecolor{darkgreen}{rgb}{0,0.7,0}
\definecolor{redcam}{RGB}{255, 95, 62}
\definecolor{bluecam}{RGB}{102, 126, 255}

\newcommand\boldp[1]{\vspace{.1cm}\noindent\textbf{#1:}\hspace{.065cm}}
\newcommand\boldpnov[1]{\noindent\textbf{#1}\hspace{.015cm}}

\begin{document}

\title{RegNeRF: Regularizing Neural Radiance Fields\\for View Synthesis from Sparse Inputs}

\author{\vspace{.05cm} Michael Niemeyer$^{1,2,3}$\thanks{The work was primarily done during an internship at Google.} \quad Jonathan T. Barron$^{3}$ \quad Ben Mildenhall$^{3}$\\ \vspace{.2cm} Mehdi S.\ M.\ Sajjadi$^{3}$ \quad Andreas Geiger$^{1,2}$ \quad Noha Radwan$^{3}$\\
\vspace{.05cm}$^1$Max Planck Institute for Intelligent Systems, Tübingen \quad $^2$University of Tübingen \\ $^3$Google Research\\
{\tt\small \{firstname.lastname\}@tue.mpg.de} \quad {\tt\small\{barron, bmild, msajjadi, noharadwan\}@google.com}
}

\maketitle

\begin{abstract}
    Neural Radiance Fields (NeRF) have emerged as a powerful representation for the task of novel view synthesis due to their simplicity and state-of-the-art performance. Though NeRF can produce photorealistic renderings of unseen viewpoints when many input views are available, its performance drops significantly when this number is reduced. We observe that the majority of artifacts in sparse input scenarios are caused by errors in the estimated scene geometry, and by divergent behavior at the start of training. We address this by regularizing the geometry and appearance of patches rendered from unobserved viewpoints, and annealing the ray sampling space during training. We additionally use a normalizing flow model to regularize the color of unobserved viewpoints. Our model outperforms not only other methods that optimize over a single scene, but in many cases also conditional models that are extensively pre-trained on large multi-view datasets.
\end{abstract}

\section{Introduction}
\label{sec:into}

Coordinate-based neural representations~\cite{Mescheder2019CVPR, Park2019CVPR, Chen2019CVPR, Michalkiewicz2019ICCV} have gained increasing popularity in the field of 3D vision.
In particular, Neural Radiance Fields (NeRF)~\cite{Mildenhall2020ECCV} have emerged as a powerful representation for the task of novel view synthesis, where the goal is to render unseen viewpoints of a scene from a given set of input images.
\begin{figure}
    \centering
    \begin{subfigure}[b]{1.\linewidth}
    \centering
    \includegraphics[width=.32\linewidth]{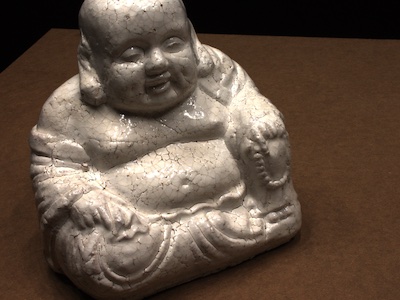}
    \includegraphics[width=.32\linewidth]{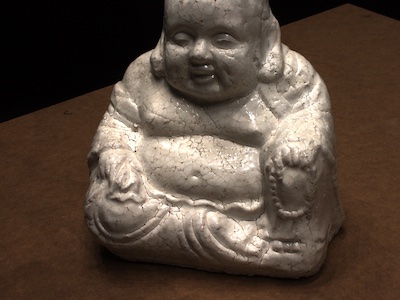}
    \includegraphics[width=.32\linewidth]{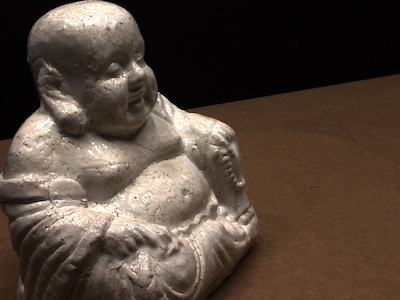}
    \caption{Sparse Set of $3$ Input Images}
    \end{subfigure}
    \begin{subfigure}[b]{1.0\linewidth}
    \centering
    \includegraphics[width=.32\linewidth]{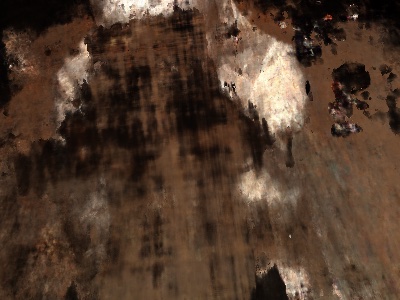}
    \includegraphics[width=.32\linewidth]{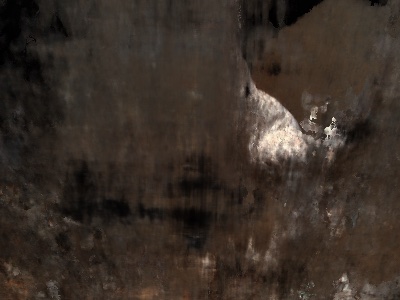}
    \includegraphics[width=.32\linewidth]{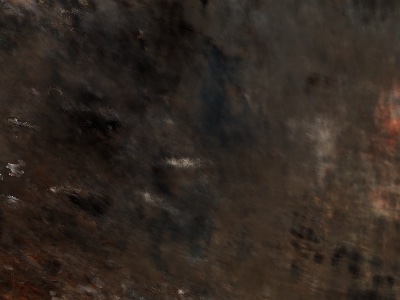}
    \caption{Novel Views Synthesized by NeRF~\cite{Mildenhall2020ECCV}}    \label{subfig:nerf}
    \end{subfigure}
    \begin{subfigure}[b]{1.0\linewidth}
    \centering
    \includegraphics[width=.32\linewidth]{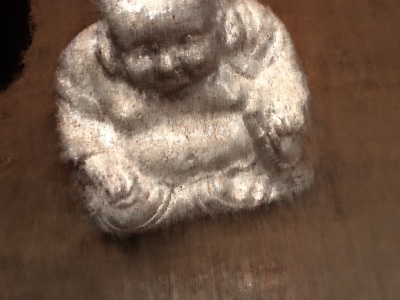}
    \includegraphics[width=.32\linewidth]{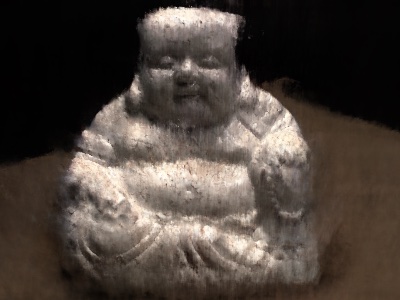}
    \includegraphics[width=.32\linewidth]{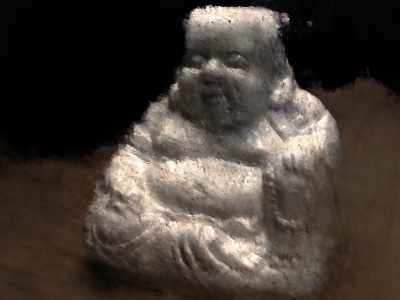}
    \caption{Same Novel Views Synthesized by Our Method}
    \label{subfig:novelviewsours}
    \end{subfigure}
    \caption{
    \textbf{View Synthesis from Sparse Inputs.}
    While Neural Radiance Fields (NeRF) allow for state-of-the-art view synthesis if many input images are provided, results degrade when only few views are available (\ref{subfig:nerf}).
    In contrast, even with sparse inputs our novel regularization and optimization strategy leads to 3D-consistent representations that render realistic novel views (\ref{subfig:novelviewsours}).
    }
    \label{fig:teaser}
    \vspace{-.3cm}
\end{figure}

Though NeRF achieves state-of-the-art performance, it requires dense coverage of the scene. However, in real-world applications such as AR/VR, autonomous driving, and robotics, the input is typically much sparser, with only few views of any particular object or region available per scene.
In this sparse setting, the quality of NeRF's rendered novel views drops significantly (see~\figref{fig:teaser}). %
\begin{figure*}
    \centering
    \includegraphics[width=1.\linewidth]{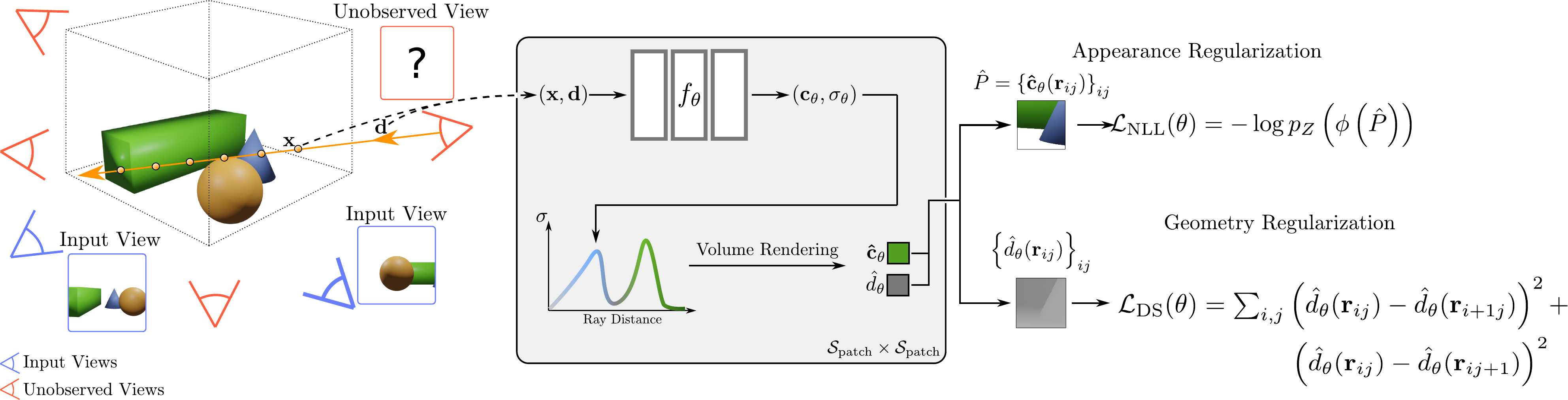}
    \vspace{-.5cm}
    \caption{
    \textbf{Overview.}
    NeRF optimizes the reconstruction loss for a given set of input images ({\color{bluecam}blue cameras}).
    For sparse inputs, however, this leads to degenerate solutions.
    In this work, we propose to sample unobserved views ({\color{redcam}red cameras}) and regularize the geometry and appearance of  patches rendered from those views. 
    More specifically, we cast rays through the scene and render patches from unobserved viewpoints for a given radiance field $f_\theta$.
    We then regularize appearance by feeding the predicted RGB patches through a trained normalizing flow model $\phi$
    and maximizing predicted log-likelihood.
    We regularize geometry by enforcing a smoothness loss on the rendered depth patches.
    Our approach leads to 3D-consistent representations even for sparse inputs from which realistic novel views can be rendered.
    }
    \label{fig:overview}
	\vspace{-0.3cm}
\end{figure*}

Several works have proposed conditional models to overcome these limitations~\cite{Yu2021CVPR, Chibane2021CVPR, Wang2021CVPRa, Chen2021ICCVa,Liu2021ARXIVa, Trevithick2021ICCV}. 
These models require expensive \textit{pre-training}, \ie training the model on large-scale datasets of many scenes with multi-view images and camera pose annotations, as opposed to \textit{test-time optimization} which is done from scratch for a given test scene.
At test time, novel views can be generated from only a few input images through amortized inference, optionally combined with per scene test time fine-tuning.
Though these models achieve promising results, obtaining the necessary pre-training data by capturing or rendering many different scenes can be prohibitively expensive.
Moreover, 
these techniques may not generalize well to novel domains at test time,
and may exhibit blurry artifacts as a result of the inherent ambiguity of sparse input data.

One alternate approach is to optimize the network weights from scratch for every new scene and introduce regularization to improve the performance for sparse inputs, \eg, by adding extra supervision~\cite{Kangle2021ARXIV} or learning embeddings representative of the input views~\cite{Jain2021ICCV}. However, existing methods either heavily rely on external supervisory signals that might not always be available, or operate on low-resolution renderings of the scene that provide only high-level information.

\boldp{Contribution} In this paper, we present \textit{RegNeRF}, a novel method for regularizing NeRF models for sparse input scenarios.
Our main contributions are the following:
\begin{compactitem} %
    \item A patch-based regularizer for depth maps rendered from unobserved viewpoints, which reduces floating artifacts and improves scene geometry.
    \item A normalizing flow model to regularize the colors predicted at unseen viewpoints by maximizing the log-likelihood of the rendered patches and thereby avoid color shifts between different views.
    \item An annealing strategy for sampling points along the ray, where we first sample scene content within a small range before expanding to the full scene bounds which prevents divergence early during training. 
\end{compactitem}

\section{Related Work}
\label{sec:rel}

\boldp{Neural Representations}%
In 3D vision, coordinate-based neural representations~\cite{Mescheder2019CVPR, Park2019CVPR, Chen2019CVPR, Michalkiewicz2019ICCV} have become a popular representation for various tasks such as 3D reconstruction~\cite{Mescheder2019CVPR, Park2019CVPR, Chen2019CVPR, Peng2020ECCV, Tancik2020NEURIPS, Niemeyer2019ICCV, Sitzmann2020NIPSb, Atzmon2019NIPS, Genova2019ICCV, Saito2019ICCVa, Gropp2020ICML, Wang2021NEURIPS, Oechsle2021ICCV}, 3D-aware generative modelling~\cite{Schwarz2020NEURIPS, Niemeyer2021CVPR, Chan2021CVPR, Niemeyer2021THREEDV, Gu2021ARXIV, Oechsle2019ICCV, Devries2021ICCV, Zhou2021ARXIV, Meng2021ICCV, Hao2021ICCV}, and novel-view synthesis~\cite{Mildenhall2020ECCV, Barron2021ICCV, Niemeyer2020CVPR, Sitzmann2019NIPS, Yariv2020NIPS, Yariv2021ARXIV, Martin-Brualla2021CVPR, Liu2020NEURIPS, Jiang2020CVPRa, Park2020CVPR, Kellnhofer2021CVPR, Bergman2021NEURIPS, Gao2020ARXIV}.
In contrast to traditional representations like point clouds, meshes, or voxels, this paradigm represents 3D geometry and color information in the weights of a neural network, leading to a compact representation. 
Several works~\cite{Niemeyer2020CVPR, Sitzmann2019NIPS, Yariv2020NIPS, Mildenhall2020ECCV, Liu2019NIPSb} proposed differentiable rendering approaches to learn neural representations from only multi-view image supervision. Among these, Neural Radiance Fields (NeRF)~\cite{Mildenhall2020ECCV} have emerged as a powerful method for novel-view synthesis due to its simplicity and state-of-the-art performance. In mip-NeRF~\cite{Barron2021ICCV}, point-based ray tracing is replaced using cone tracing to combat aliasing.
As this is a more robust representation for scenes with various camera distances and reduces NeRF's coarse and fine MLP networks to a single multiscale MLP, we adopt mip-NeRF as our scene representation.
However, compared to previous works~\cite{Barron2021ICCV, Mildenhall2020ECCV}, we consider a much sparser input scenario in which neither NeRF nor mip-NeRF are able to produce realistic novel views. 
By regularizing scene geometry and appearance, we are able to synthesize high-quality renderings despite only using as few as 3 wide-baseline input images.

\boldp{Sparse Input Novel-View Synthesis}
One approach for circumventing the requirement of dense inputs is to aggregate prior knowledge by pre-training a conditional model of radiance fields~\cite{Yu2021CVPR, Chibane2021CVPR, Wang2021CVPRa, Chen2021ICCVa, Liu2021ARXIVa, Trevithick2021ICCV, Rematas2021ICML, Li2021ICCVa, Jang2021ICCV}.
PixelNeRF~\cite{Yu2021CVPR} and Stereo Radiance Fields~\cite{Chibane2021CVPR} use local CNN features extracted from the input images, whereas MVSNeRF~\cite{Chen2021ICCVa} obtains a 3D cost volume via image warping which is then processed by a 3D CNN. 
Though they achieve compelling results, these methods require a multi-view image dataset of many different scenes for pre-training, which is not always readily available and may be expensive to obtain.
Further, most approaches require fine-tuning the network weights at test time despite the long pre-training phase, and the quality of novel views is prone to drop when the data domain changes at test time.
Tancik~\etal \cite{Tancik2021CVPR} learn network initializations from which test time optimization on a new scene converges faster. This approach assumes that the training and test data are taken from the same domain, and results may degrade if the domain changes at test time.

In this work, we explore an alternative approach which avoids expensive pre-training by regularizing appearance and geometry in novel (virtual) views.
Previous works in this direction include DS-NeRF~\cite{Kangle2021ARXIV} and DietNeRF~\cite{Jain2021ICCV}.
DS-NeRF improves reconstruction accuracy by adding additional depth supervision. In contrast, our approach only uses RGB images and does not require depth input.
DietNeRF~\cite{Jain2021ICCV} compares CLIP~\cite{Radford2021ICML, Dosovitskiy2021ICLR} embeddings of unseen viewpoints rendered at low resolutions. This semantic consistency loss can only provide high-level information and does not improve scene geometry for sparse inputs.
Our approach instead regularizes scene geometry and appearance based on rendered patches and applies a scene space annealing strategy. We find that our approach leads to more realistic scene geometry and more accurate novel views. 

\section{Method}
\label{sec:method}

We propose a novel optimization procedure for neural radiance fields from sparse inputs.
More specifically, our approach builds upon mip-NeRF~\cite{Barron2021ICCV}, which uses a multi-scale radiance field model to represent scenes (\secref{subsec:background}).
For sparse views, we find the quality of mip-NeRF's view synthesis drops mainly due to incorrect scene geometry and training divergence. 
To overcome this, we propose a patch-based approach to regularize the predicted color and geometry from unseen viewpoints (\secref{subsec:patch-reg}).
We also provide a strategy for annealing the scene sampling bounds to avoid divergence at the beginning of training (\secref{subsec:sample-anneal}).
Finally, we use higher learning rates in combination with gradient clipping to speed up the optimization process (\secref{subsec:details}). 
\figref{fig:overview} shows an overview of our method.

\subsection{Background}
\label{subsec:background}

\boldpnov{Neural Radiance Fields} A radiance field is a continuous function $f$ mapping a 3D location $\bx \in \nR^3$ and viewing direction $\bd \in \nS^2$ to a volume density $\sigma \in [0, \infty)$ and color value $\bc \in [0,1]^3$. Mildenhall~\etal~\cite{Mildenhall2020ECCV} parameterize this function using a multi-layer perceptron (MLP), where the weights of the MLP are optimized to reconstruct a set of input images of a particular scene:
\begin{align}
\begin{split}
    f_\theta: \nR^{L_\bx} \times \nR^{L_\bd} &\to [0,1]^3 \times [0, \infty) \\
    \left(\gamma(\bx), \gamma(\bd)\right) &\mapsto (\bc, \sigma)\,.
\end{split}
\end{align}
Here, $\theta$ indicates the network weights and $\gamma$ a predefined positional encoding~\cite{Mildenhall2020ECCV, Tancik2020NEURIPS} applied to $\bx$ and $\bd$.

\boldp{Volume Rendering} Given a neural radiance field $f_\theta$, a pixel is rendered by casting a ray $\br(t) = \bo + t\bd$ from the camera center $\bo$ through the pixel along direction $\bd$.
For given near and far bounds $t_n$ and $t_f$, the pixel's predicted color value $\hat\bc_\theta$ is computed using alpha compositing:
\begin{align}
    \begin{split}\label{eq:color-rendering}
        \hat\bc_\theta(\br) = \int_{t_n}^{t_f} T(t) \sigma_\theta(\br(t)) \bc_\theta(\br(t), \bd) \,dt \\
        \text{where}\quad T(t) = \exp \left(-\int_{t_n}^{t} \sigma_\theta(\br(s)) \,ds \right)\,,
    \end{split}
\end{align}
and $\sigma_\theta(\cdot)$ and $\bc_\theta(\cdot, \cdot)$ indicate the density and color prediction of radiance field $f_\theta$, respectively. In practice, these integrals are approximated using quadrature~\cite{Mildenhall2020ECCV}. A neural radiance field is optimized over a set of input images and their camera poses by minimizing the mean squared error
\begin{align}\label{eq:mse}
    \cL_\text{MSE}(\theta, \cR_i) = \sum_{\br \in \cR_i} \left\| \hat\bc_\theta(\br) - \bc_\text{GT}(\br)  \right\|^2\,,
\end{align}
where $\cR_i$ indicates a set of input rays and $\bc_\text{GT}$ its GT color.

\boldp{mip-NeRF} While NeRF only casts a single ray per pixel, mip-NeRF~\cite{Barron2021ICCV} instead casts a cone. The positional encoding changes from representing an infinitesimal point to an integration over a volume covered by a conical frustum.
This is a more appropriate representation for scenes with varying camera distances and allows NeRF's coarse and fine MLPs to be combined into a single multiscale MLP, thereby increasing training speed and reducing model size.
We adopt the mip-NeRF representation in this work.

\subsection{Patch-based Regularization}
\label{subsec:patch-reg}

NeRF's performance drops significantly if the number of input views is sparse.
Why is this the case?
Analyzing its optimization procedure, the model is only supervised from these sparse viewpoints by the reconstruction loss in~\eqref{eq:mse}. 
While it learns to reconstruct the input views perfectly, novel views may be degenerate because the model is not biased towards learning a 3D consistent solution in such a sparse input scenario (see~\figref{fig:teaser}).
To overcome this limitation, we regularize unseen viewpoints. More specifically, we define a space of unseen but relevant viewpoints and render small patches randomly sampled from these cameras.
Our key idea is that these patches can be regularized to yield smooth geometry and high-likelihood colors.

\boldp{Unobserved Viewpoint Selection}
To apply regularization techniques for unobserved viewpoints, we must first define the sample space of unobserved camera poses.
We assume a known set of target poses $\left\{ \bP^i_\text{target} \right\}_i$ where
\begin{align}
    \bP_\text{target}^i = \left[ \bR^i_\text{target} | \bt^i_\text{target} \right] \in SE(3)\,.
\end{align}
These target poses can be thought of bounding the set of poses from which we would like to render novel views at test time.
We define the space of possible camera locations as the bounding box of all given target camera locations
\begin{align}
    \cS_t = \left\{ \bt \in \nR^3 \mid \bt_\text{min} \leq \bt \leq \bt_\text{max} \right\}
\end{align}
where $\bt_\text{min}$ and $\bt_\text{max}$ are the elementwise minimum and maximum values of $\left\{ \bt^i_\text{target} \right\}_i$, respectively.

To obtain the sample space of camera rotations, we assume that all cameras roughly focus on a central scene point. We define a common ``up'' axis $\bar\bp_u$ by computing the normalized mean over the up axes of all target poses. Next, we calculate a mean focus point $\bar\bp_f$ by solving a least squares problem to determine the 3D point with minimum squared distance to the optical axes of all target poses.
To learn more robust representations, we add random jitter to the focal point before calculating the camera rotation matrix. 
We define the set of of all possible camera rotations (given the sampled position $\bt$) as
\begin{align}
    \cS_R | \bt = \left\{\bR(\bar\bp_u, \bar\bp_f + \mathbf{\epsilon}, \bt) \mid \mathbf{\mathbf{\epsilon}} \sim \cN(0, 0.125) \right\}
\end{align}
where $\bR(\cdot, \cdot, \cdot)$ indicates the resulting ``look-at'' camera rotation matrix and $\epsilon$ is a small jitter added to the focus point.
We obtain a random camera pose by sampling a position and rotation:
\begin{align}
    \cS_{P} = \left\{ \left[ \bR \vert \bt \right] | \bR \sim \cS_R|\bt, \bt \sim \cS_t \right\}
\end{align}

\boldp{Geometry Regularization} It is well-known that real-world geometry tends to be piece-wise smooth, \ie, flat surfaces are more likely than high-frequency structures~\cite{Huang2000CVPR}. 
We incorporate this prior into our model by encouraging depth smoothness from unobserved viewpoints.
Similarly to how a pixel's color is rendered in~\eqref{eq:color-rendering}, we calculate the expected depth as:
\begin{align}
    \begin{split}\label{eq:depth-rendering}
        \hat{d_\theta}(\br) = \int_{t_n}^{t_f} T(t) \sigma_\theta(\br(t)) t \,dt \,.
    \end{split}
\end{align}
We formulate our depth smoothness loss as
\begin{align}\begin{split}
    \cL_\text{DS}(\theta, \cR_r) = \sum_{\br \in \cR_r} \sum_{i, j=1}^{S_\text{patch} - 1}  & \left( \hat{d_\theta}(\br_{ij}) - \hat{d_\theta}(\br_{i+1j}) \right)^2 \\
     + & \left( \hat{d_\theta}(\br_{ij}) - \hat{d_\theta}(\br_{ij+1}) \right)^2\,,
\end{split}\end{align}
where $\cR_r$ indicates a set of rays sampled from camera poses $\cS_{P}$, $\br_{ij}$ is the ray through pixel $(i, j)$ of a patch centered at $\br$, and $S_\text{patch}$ is the size of the rendered patches.

\boldp{Color Regularization}
We observe that for sparse inputs, the majority of artifacts are caused by incorrect scene geometry.
However, even with correct geometry, optimizing a NeRF model can still lead to color shifts or other errors in scene appearance prediction due to the sparsity of the inputs. 
To avoid degenerate colors and ensure stable optimization, we also regularize color prediction.
Our key idea is to estimate the likelihood of rendered patches and maximize it during optimization. To this end, we make use of readily-available unstructured 2D image datasets. Note that, while datasets of posed multi-view images are expensive to collect, collections of unstructured natural images are abundant.
Our only criterion for the dataset is that it contains diverse natural images, allowing us to reuse the same flow model for any type of real-world scene we reconstruct.
We train a RealNVP~\cite{Dinh2017ICLR} normalizing flow model on patches from the JFT-300M~dataset~\cite{Sun2017ICCVa}. 
With this trained flow model we estimate the log-likelihoods (LL) of rendered patches and maximize them during optimization. Let 
\begin{align}
\phi: [0,1]^{S_\text{patch} \times S_\text{patch} \times 3} \to \nR^d
\end{align}
be the learned bijection mapping an RGB patch of size $S_\text{patch}=8$ 
to $\nR^d$ where $d = S_\text{patch} \cdot S_\text{patch} \cdot 3$. We define our color regularization loss as
\begin{align}
    \begin{split}
    \cL_\text{NLL}(\theta, \cR_r) = \sum_{\br \in \cR_r} -\log p_{Z} \left( \phi\left( \hat{P}_r \right)\right)\\
    \text{where } \hat{P}_r = \left\{ \hat\bc_\theta (\br_{ij}) \mid 1 \leq i, j \leq S_\text{patch} \right\}
    \end{split}
\end{align}
and $\cR_r$ indicates a set of rays sampled from $\cS_{P}$, $\hat{P}_r$ the predicted RGB color patch with center $\br$, and $-\log p_Z$ the negative log-likelihood (``NLL'') with Gaussian $p_Z$.

\boldp{Total Loss} The total loss we optimize in each iteration is
\begin{align}
    \cL(\theta) =  \cL_\text{MSE}(\theta, \cR_i) + \cL_\text{DS}(\theta, \cR_r) + \cL_\text{NLL}(\theta, \cR_r)
\end{align}
where $\cR_i$ indicates a set of rays from input poses, and $\cR_r$ a set of rays from random poses $\cS_{P}$.

\subsection{Sample Space Annealing}
\label{subsec:sample-anneal}

For very sparse scenarios (\eg, 3 or 6 input views), we observe another failure mode of NeRF: divergent behavior at the start of training. 
This leads to high density values at ray origins. While input views are correctly reconstructed, novel views degenerate as no 3D-consistent representation is recovered. 
We find that annealing
the sampled scene space quickly over the early iterations during optimization helps to avoid this problem.
By restricting the scene sampling space to a smaller region defined for all input images, we introduce an inductive bias to explain the input images with geometric structure in the center of the scene. %

Recall from~\eqref{eq:color-rendering} that $t_n, t_f$ are the camera's near and far plane, respectively, and let $t_m$ be a defined center point (usually the midpoint between $t_n$ and $t_f$).
We define
\begin{align}
\begin{split}
    t_n(i) &= t_m + (t_n - t_m) \eta(i) \\
    t_f(i) &= t_m + (t_f - t_m) \eta(i) \\
    \eta(i) &= \mathrm{max}\left(\mathrm{min}\left( \sfrac{i}{N_t},\, p_s \right), \, 1\right)
    \end{split}
\end{align}
    
where $i$ indicates the current training iteration, $N_t$ a hyperparameter indicating how many iterations until the full range is reached, and $p_s$ a hyperparameter indicating a start range (\eg, $0.5$).
This annealing is applied to renderings from both the input poses and the sampled unobserved viewpoints.
We find that this annealing strategy ensures stability during early training and avoids degenerate solutions.

\subsection{Training Details}
\label{subsec:details}

We build our code on top of of the JAX~\cite{jax2018github} mip-NeRF codebase.\footnote{Official codebase available at \url{https://github.com/google/mipnerf}.} 
We optimize with Adam~\cite{Kingma2015ICLR} using an exponential learning rate decay from $2 \cdot 10^{-3}$ to $2 \cdot 10^{-5}$.
We clip gradients by value at $0.1$ and then by norm at $0.1$. 
We train for \num{500} pixel epochs, %
\eg, \num{44}K, \num{88}K, and \num{132}K iterations on DTU for $3 / 6 / 9$ input views respectively
(all fewer iterations than mip-NeRF's default \num{250}K steps~\cite{Mildenhall2020ECCV}).

\section{Experiments}
\label{sec:experiments}
\begin{figure}
\captionsetup[subfigure]{justification=centering}
    \centering
     \begin{subfigure}[b]{1.0\linewidth}
         \centering
        \includegraphics[width=.32\linewidth]{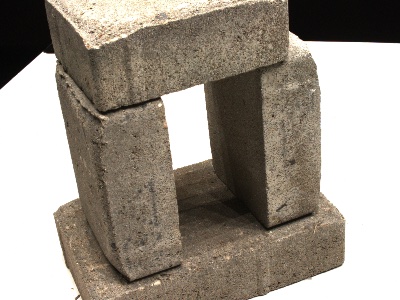}
        \includegraphics[width=.32\linewidth]{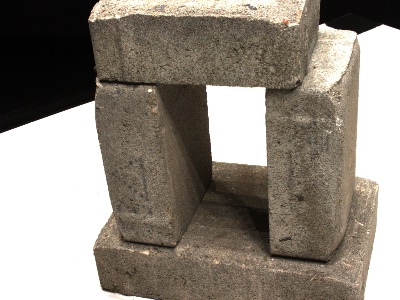}
        \includegraphics[width=.32\linewidth]{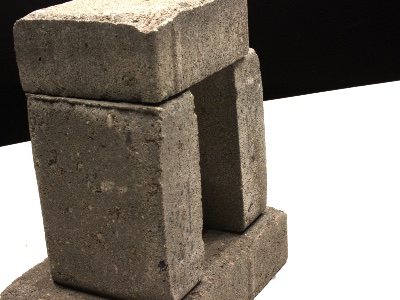}
        \caption{Sparse Set of 3 Input Views}\label{subfig:dtueval-input}
     \end{subfigure}
     \begin{subfigure}[t]{.32\linewidth}
         \centering
        \includegraphics[width=\linewidth]{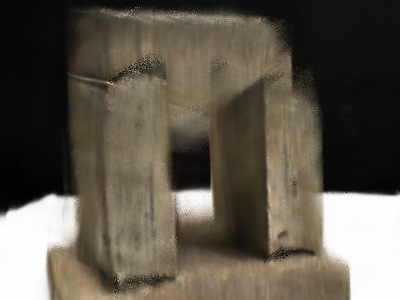}
        \caption{PixelNeRF\\(PSNR: $17.37 / 16.89$)}\label{subfig:dtueval-pn}
     \end{subfigure}
     \begin{subfigure}[t]{.32\linewidth}
         \centering
        \includegraphics[width=\linewidth]{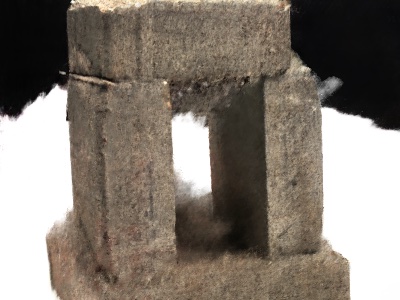}
         \caption{Our Method\\(PSNR: $8.79 / 20.08$)}\label{subfig:dtueval-ours}
     \end{subfigure}
     \begin{subfigure}[t]{.32\linewidth}
         \centering
        \includegraphics[width=\linewidth]{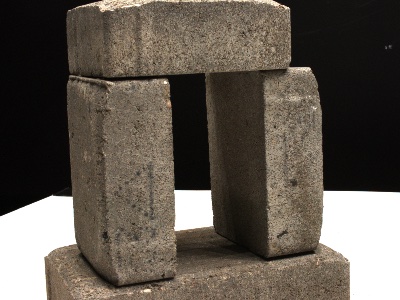}
    \caption{GT}\label{subfig:dtueval-gt}
     \end{subfigure}
     \vspace{-.1cm}
    \caption{
    \textbf{Evaluation Bias.}
        Many scenes in DTU are composed of an object on a white table with black background resulting in an evaluation bias favoring a correct background over the object-of-interest.
        For sparse inputs the background may only be partially observed in the input views and strongly overfitting to the table is incentivized,
        though most real-world applications would prefer to accurately reconstruct the object-of-interest.
        Here we show an example with full image (first) and object-of-interest (second) PSNR for PixelNeRF (which drops from $17.37$ to $16.89$) and for our method (which improves from $8.79$ to $20.08$).
    }
    \label{fig:dtu-eval-bias}
    \vspace{-.5cm}
\end{figure}
     
\boldpnov{Datasets} We report results on the real world multi-view datasets DTU~\cite{Jensen2014CVPR} and LLFF~\cite{Mildenhall2019SIGGRAPH}. 
DTU contains images of objects placed on a table, and LLFF consists of complex forward-facing scenes. For DTU, we observe that in scenes with a white table and a black background, the model is heavily penalized for incorrect background predictions regardless of the quality of the rendered object-of-interest (see~\figref{fig:dtu-eval-bias}).
To avoid this background bias, we evaluate all methods with the object masks applied to the rendered images (full image evaluations in supp.\ mat.).
We adhere to the protocol of Yu \etal~\cite{Yu2021CVPR} and evaluate on their reported test set of $15$ scenes. For LLFF, we adhere to community standards~\cite{Mildenhall2020ECCV} and use every $8$-th image as the held-out test set and select the input views evenly from the remaining images. Following previous work~\cite{Yu2021CVPR}, we report results for the scenarios of $3$, $6$, and $9$ input views.

\boldp{Metrics} We report the mean of PSNR, structural similarity index (SSIM)~\cite{Wang2004TIP}, and the LPIPS perceptual metric ~\cite{Zhang2018CVPRa}.
To ease comparison, we also report the geometric mean of $\text{MSE} = 10^{-\text{PSNR}/10}$, $\sqrt{1 - \text{SSIM}}$, and LPIPS~\cite{Barron2021ICCV}.

\begin{figure}
    \centering
    \begin{tabular}{P{.199\linewidth}P{.199\linewidth}P{.199\linewidth}P{.199\linewidth}}
         \scriptsize mip-NeRF~\cite{Barron2021ICCV} & \scriptsize Ours & \scriptsize mip-NeRF~\cite{Barron2021ICCV} & \scriptsize Ours\\
    \end{tabular}
    \begin{subfigure}[b]{1.0\linewidth}
         \centering
        \includegraphics[width=1.\linewidth]{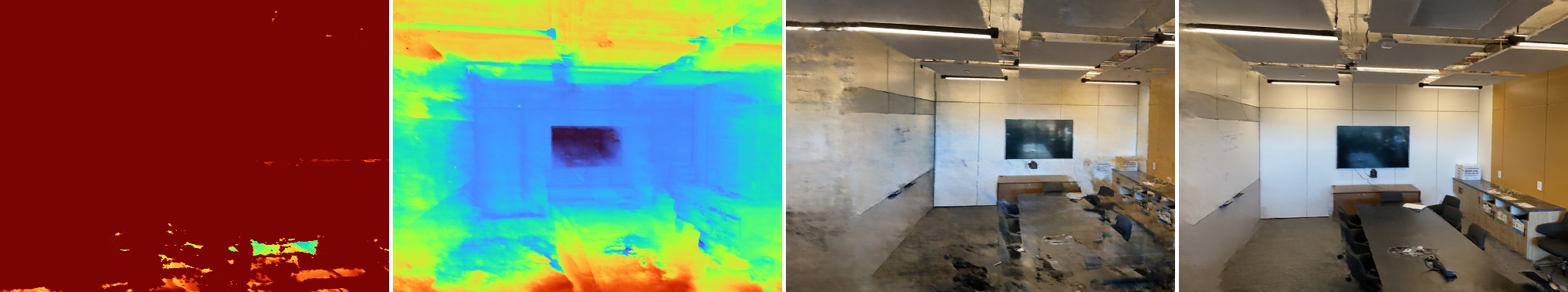}
         \caption{3 Input Views}
         \vspace{.125cm}
     \end{subfigure}
    \begin{subfigure}[b]{1.0\linewidth}
         \centering
        \includegraphics[width=1.\linewidth]{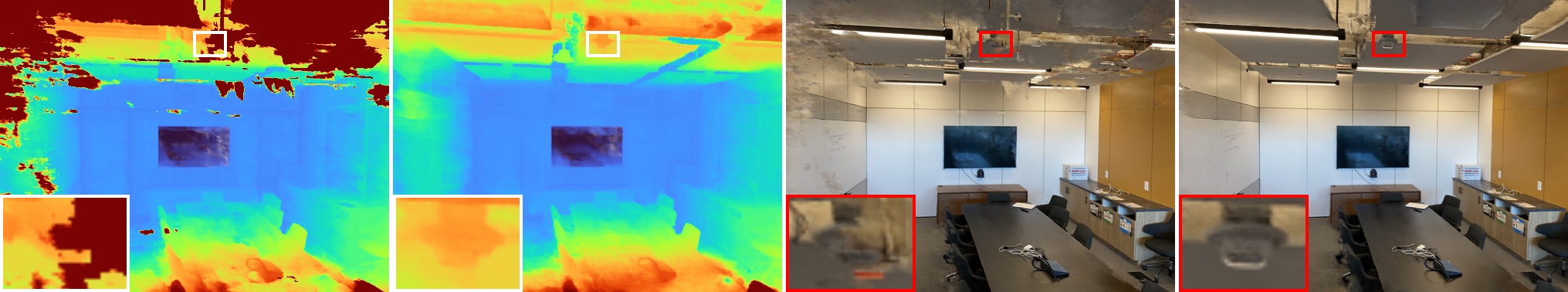}
         \caption{6 Input Views}
         \vspace{.125cm}
     \end{subfigure}
    \begin{subfigure}[b]{1.0\linewidth}
         \centering
        \includegraphics[width=1.\linewidth]{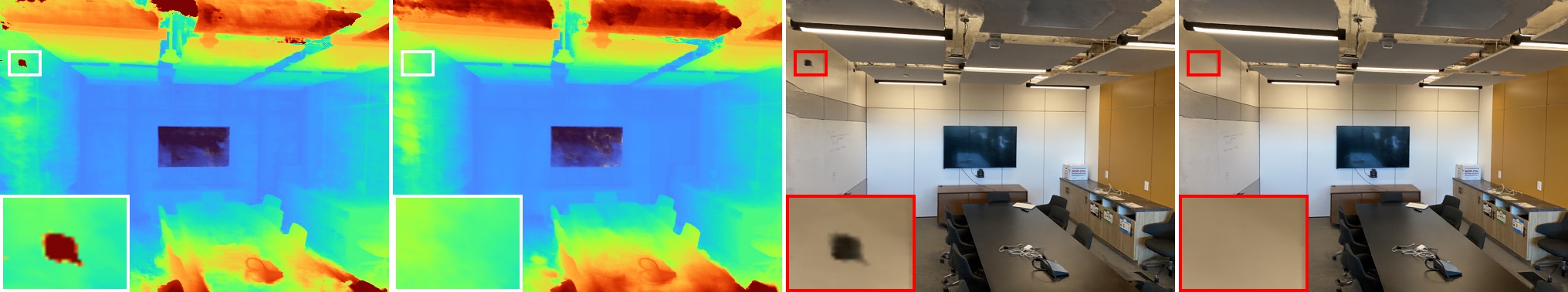}
         \caption{9 Input Views}
     \end{subfigure}
     \vspace{-.6cm}
    \caption{
        \textbf{Importance of Geometry.}
        We compare expected depth maps (left) and RGB renderings (right) for mip-NeRF~\cite{Barron2021ICCV} and our method on the LLFF dataset.
        The quality of optimized geometry is  correlated with view synthesis performance: Our proposed scene space annealing and geometry regularization strategies remove floating artifacts (see zoom-in) and lead to smooth geometry, which in turn leads to improved quality of rendered novel views. 
        }
    \label{fig:geometry}
\end{figure}
\boldp{Baselines}
We compare against the state-of-the-art conditional models PixelNeRF~\cite{Yu2021CVPR}, Stereo Radiance Fields (SRF)~\cite{Chibane2021CVPR}, and MVSNeRF~\cite{Chen2021ICCVa}.
We re-train PixelNeRF for the $6 / 9$ view scenarios, leading to better results, and we similarly pre-train SRF with $3 / 6 / 9$ views.
We pre-train all methods on the large-scale DTU~dataset.
The LLFF dataset has been shown to be too small for pre-training~\cite{Kangle2021ARXIV} and hence serves as an out-of-distribution test for conditional models.
We report the conditional models on both datasets also after additional per-scene test time optimization (``ft'' for ``fine-tuned'').
Further, we compare against mip-NeRF~\cite{Barron2021ICCV} and DietNeRF~\cite{Jain2021ICCV} which do not require pre-training, as with our approach. As no official code is available, we reimplement DietNeRF on top of the mip-NeRF codebase (achieving better results) and train both methods for \num{250}K iterations per scene with exponential learning rate decay from $5 \cdot 10^{-4}$ to $5 \cdot 10^{-5}$.

\subsection{View Synthesis from Sparse Inputs}
We first compare our model to the vanilla mip-NeRF baseline, analyzing the effect of our regularizers on scene geometry, appearance and data efficiency.
\begin{table*}
    \resizebox{\linewidth}{!}{\begin{tabular}{l|c|ccc|ccc|ccc|ccc}
\toprule
  & \multirow{2}{*}{Setting} &  \multicolumn{3}{c}{PSNR $\uparrow$} & \multicolumn{3}{c}{SSIM $\uparrow$} & \multicolumn{3}{c}{LPIPS $\downarrow$} & \multicolumn{3}{c}{Average $\downarrow$}  \\
  &  & 3-view & 6-view & 9-view  & 3-view & 6-view & 9-view  & 3-view & 6-view & 9-view  & 3-view & 6-view & 9-view \\ \midrule
SRF~\cite{Chibane2021CVPR} & \multirow{3}{*}{Trained on DTU} & 15.32 & 17.54 & 18.35 & 0.671 & 0.730 & 0.752 & 0.304 & 0.250 & 0.232 & 0.171 & 0.132 & 0.120 \\
PixelNeRF~\cite{Yu2021CVPR} &  & 16.82 & 19.11 & 20.40 & 0.695 & 0.745 & 0.768 & 0.270 & 0.232 & 0.220 & 0.147 & 0.115 & 0.100 \\
MVSNeRF~\cite{Chen2021ICCVa} &  &  \cellcolor{yellow!25}18.63 &  \cellcolor{orange!25}20.70 & 22.40 &  \cellcolor{red!25}0.769 &  \cellcolor{orange!25}0.823 &  \cellcolor{yellow!25}0.853 &  \cellcolor{orange!25}0.197 &  \cellcolor{yellow!25}0.156 &  \cellcolor{yellow!25}0.135 &  \cellcolor{orange!25}0.113 &  \cellcolor{orange!25}0.088 & 0.068 \\
\midrule 
SRF ft~\cite{Chibane2021CVPR} & \multirow{3}{*}{\shortstack{Trained on DTU\\and\\Optimized per Scene}} & 15.68 & 18.87 & 20.75 & 0.698 & 0.757 & 0.785 & 0.281 & 0.225 & 0.205 & 0.162 & 0.114 & 0.093 \\
PixelNeRF ft~\cite{Yu2021CVPR} &  &  \cellcolor{red!25}18.95 & 20.56 & 21.83 & 0.710 & 0.753 & 0.781 & 0.269 & 0.223 & 0.203 & 0.125 & 0.104 & 0.090 \\
MVSNeRF ft~\cite{Chen2021ICCVa} &  & 18.54 & 20.49 & 22.22 &  \cellcolor{red!25}0.769 &  \cellcolor{yellow!25}0.822 &  \cellcolor{yellow!25}0.853 &  \cellcolor{orange!25}0.197 &  \cellcolor{orange!25}0.155 &  \cellcolor{yellow!25}0.135 &  \cellcolor{orange!25}0.113 &  \cellcolor{yellow!25}0.089 & 0.069 \\
\midrule 
mip-NeRF~\cite{Barron2021ICCV} & \multirow{3}{*}{Optimized per Scene} & 8.68 & 16.54 &  \cellcolor{yellow!25}23.58 & 0.571 & 0.741 &   \cellcolor{orange!25}0.879 & 0.353 & 0.198 &  \cellcolor{orange!25}0.092 & 0.323 & 0.148 &  \cellcolor{orange!25}0.056 \\
DietNeRF~\cite{Jain2021ICCV} &  & 11.85 &  \cellcolor{yellow!25}20.63 &  \cellcolor{orange!25}23.83 & 0.633 & 0.778 & 0.823 & 0.314 & 0.201 & 0.173 & 0.243 & 0.101 &  \cellcolor{yellow!25}0.068 \\
\textbf{Ours} &  & \cellcolor{orange!25}18.89 &  \cellcolor{red!25}22.20 &  \cellcolor{red!25}24.93 &  \cellcolor{yellow!25}0.745 &  \cellcolor{red!25}0.841 &  \cellcolor{red!25}0.884 &  \cellcolor{red!25}0.190 &  \cellcolor{red!25}0.117 &  \cellcolor{red!25}0.089 &  \cellcolor{red!25}0.112 &  \cellcolor{red!25}0.071 &  \cellcolor{red!25}0.047 \\
\bottomrule
\end{tabular}
}
    \caption{
    \textbf{Quantitative Comparison on DTU.}
    For $3$ input views, our model achieves quantitative results comparable to conditional models (SRF, PixelNeRF, MVSNeRF) despite not requiring an expensive pre-training phase, and strongly outperforms other baselines (mip-NeRF, DietNeRF) which operate in the same setting as us.
    For $6$ and $9$ input views, our model achieves the best overall quantitative results.
    }
    \label{tab:dtu}
\end{table*}
\begin{table*}
    \resizebox{\linewidth}{!}{
\begin{tabular}{l|c|ccc|ccc|ccc|ccc}
\toprule
  & \multirow{2}{*}{Setting} &  \multicolumn{3}{c}{PSNR $\uparrow$} & \multicolumn{3}{c}{SSIM $\uparrow$} & \multicolumn{3}{c}{LPIPS $\downarrow$} & \multicolumn{3}{c}{Average $\downarrow$}  \\
  &  & 3-view & 6-view & 9-view  & 3-view & 6-view & 9-view  & 3-view & 6-view & 9-view  & 3-view & 6-view & 9-view \\ \midrule
SRF~\cite{Chibane2021CVPR} & \multirow{3}{*}{Trained on DTU} & 12.34 & 13.10 & 13.00 & 0.250 & 0.293 & 0.297 & 0.591 & 0.594 & 0.605 & 0.313 & 0.293 & 0.296 \\
PixelNeRF~\cite{Yu2021CVPR} &  & 7.93 & 8.74 & 8.61 & 0.272 & 0.280 & 0.274 & 0.682 & 0.676 & 0.665 & 0.461 & 0.433 & 0.432 \\
MVSNeRF~\cite{Chen2021ICCVa} &  & \cellcolor{yellow!25}17.25 & 19.79 & 20.47 & \cellcolor{yellow!25}0.557 & 0.656 & 0.689 & \cellcolor{yellow!25}0.356 & 0.269 & 0.242 & \cellcolor{yellow!25}0.171 & 0.125 & 0.111 \\
\midrule 
SRF ft~\cite{Chibane2021CVPR} & \multirow{3}{*}{\shortstack{Trained on DTU\\and\\Optimized per Scene}} & 17.07 & 16.75 & 17.39 & 0.436 & 0.438 & 0.465 & 0.529 & 0.521 & 0.503 & 0.203 & 0.207 & 0.193 \\
PixelNeRF ft~\cite{Yu2021CVPR} &  & 16.17 & 17.03 & 18.92 & 0.438 & 0.473 & 0.535 & 0.512 & 0.477 & 0.430 & 0.217 & 0.196 & 0.163 \\
MVSNeRF ft~\cite{Chen2021ICCVa} &  & \cellcolor{orange!25}17.88 & 19.99 & 20.47 & \cellcolor{orange!25}0.584 & 0.660 & 0.695 & \cellcolor{red!25}0.327 & 0.264 & 0.244 & \cellcolor{orange!25}0.157 & 0.122 & 0.111 \\
\midrule 
mip-NeRF~\cite{Barron2021ICCV} & \multirow{3}{*}{Optimized per Scene} & 14.62 & \cellcolor{yellow!25}20.87 & \cellcolor{yellow!25}24.26 & 0.351 &\cellcolor{yellow!25}0.692 & \cellcolor{orange!25}0.805 & 0.495 & \cellcolor{yellow!25}0.255 & \cellcolor{orange!25}0.172 & 0.246 & \cellcolor{yellow!25}0.114 & \cellcolor{orange!25}0.073 \\
DietNeRF~\cite{Jain2021ICCV} &  & 14.94 & \cellcolor{orange!25}21.75 & \cellcolor{orange!25}24.28 & 0.370 & \cellcolor{orange!25}0.717 & \cellcolor{yellow!25}0.801 & 0.496 & \cellcolor{orange!25}0.248 & \cellcolor{yellow!25}0.183 & 0.240 & \cellcolor{orange!25}0.105 & \cellcolor{orange!25}0.073 \\
\textbf{Ours} &  & \cellcolor{red!25}19.08 & \cellcolor{red!25}23.10 & \cellcolor{red!25}24.86 & \cellcolor{red!25}0.587 & \cellcolor{red!25}0.760 & \cellcolor{red!25}0.820 & \cellcolor{orange!25}0.336 & \cellcolor{red!25}0.206 & \cellcolor{red!25}0.161 & \cellcolor{red!25}0.146 & \cellcolor{red!25}0.086 & \cellcolor{red!25}0.067 \\
\bottomrule
\end{tabular}
}
    \caption{
    \textbf{Quantitative Comparison on LLFF.}
    Some conditional models (SRF, PixelNeRF) overfit to the training data (DTU) but all benefit from additional fine-tuning at test time.
    The two unconditional baselines mip-NeRF and DietNeRF do not achieve competitive results for $3$ input views, but outperform conditional models for the $6 / 9$ input view scenarios. Our method achieves the best results for all scenarios. 
    }
    \label{tab:llff}
\end{table*}

\boldp{Geometry Prediction}
We observe that novel view synthesis performance is directly correlated with how accurately the scene geometry is predicted: in~\figref{fig:geometry} we show expected depth maps and RGB renderings for mip-NeRF and our method on the LLFF room scene.
We find that for $3$ input views, mip-NeRF produces low-quality renderings and poor geometry. In contrast, our method produces an acceptable novel view and a realistic scene geometry, despite the low number of inputs. 
When increasing the number of input images to $6$ or $9$, mip-NeRF's predicted geometry improves but still contains floating artifacts. %
Our method generates smooth scene geometry, which is reflected in its higher-quality novel views.

\boldp{Data Efficiency}
\begin{figure}
    \centering
    \includegraphics[width=\linewidth]{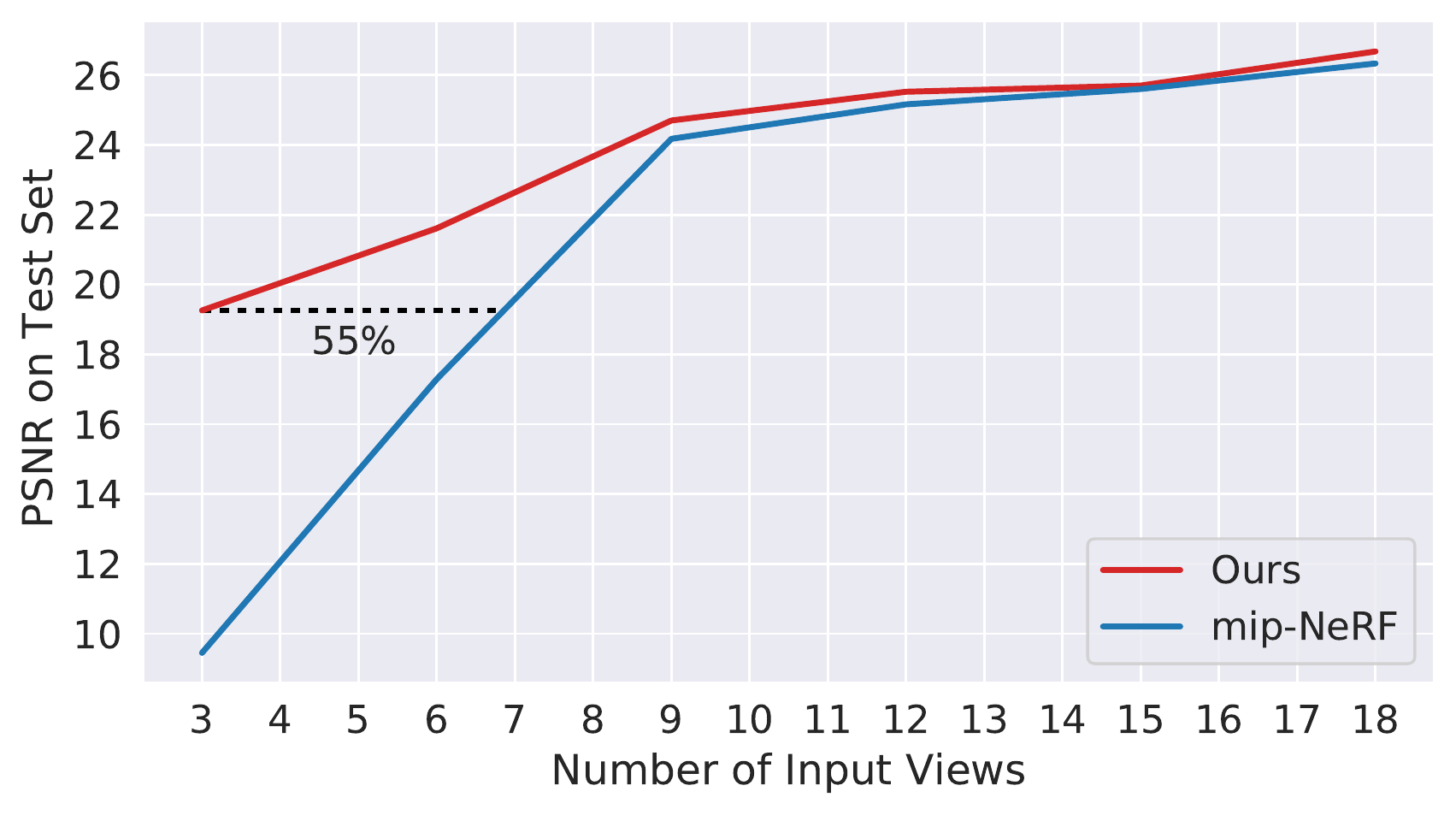}
 \vspace{-.6cm}
    \caption{
        \textbf{Data Efficiency.}
        In sparse settings, our method requires up to $55\%$ fewer images than mip-NeRF~\cite{Barron2021ICCV} to achieve a similar test set performance on the DTU dataset.
        }
    \label{fig:mipnerf_match}
    \vspace{-.65cm}
\end{figure}
To evaluate our gain in data efficiency, we train mip-NeRF and our method for various numbers of input views and compare their performance.\footnote{Results slightly differ from~\tabref{tab:dtu} as a smaller test set has to be used.} We find that for sparse inputs our method requires up to $55\%$ fewer input views to match mip-NeRF's mean PSNR on the test set, where the difference is larger for fewer input views.
For $18$ input views, both methods achieve a similar performance (as this work focuses on sparse inputs, tuning hyper-parameters for more input views could result in improved performance for these scenarios).

\subsection{Baseline Comparison}

\begin{figure*}
    \centering
    \begin{tabular}{P{0.175\textwidth}P{0.175\textwidth}P{0.175\textwidth}P{0.175\textwidth}P{0.175\textwidth}}
     \scriptsize PixelNeRF~\cite{Yu2021CVPR} & \scriptsize  MVSNeRF~\cite{Chen2021ICCVa} & \scriptsize  DietNeRF~\cite{Jain2021ICCV} & \scriptsize  Ours & \scriptsize  GT\\
    \end{tabular}
     \begin{subfigure}[b]{1.0\textwidth}
         \centering
        \includegraphics[width=\linewidth]{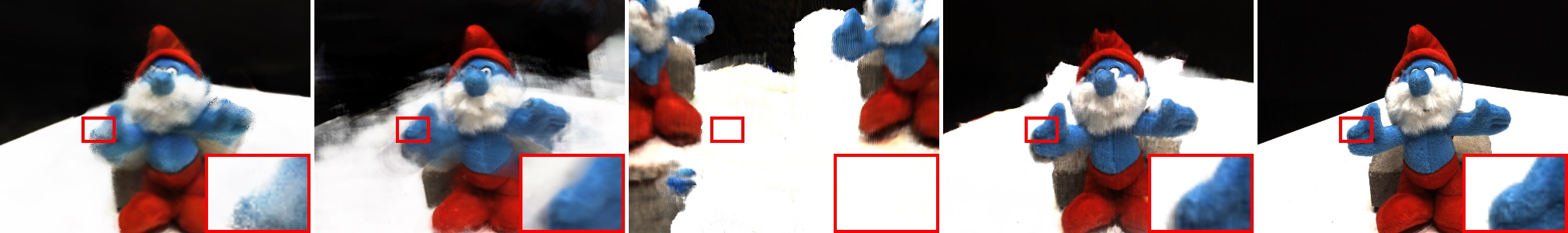}
         \caption{3 Input Views}
     \end{subfigure}
     \begin{subfigure}[b]{1.0\textwidth}
         \centering
        \includegraphics[width=\linewidth]{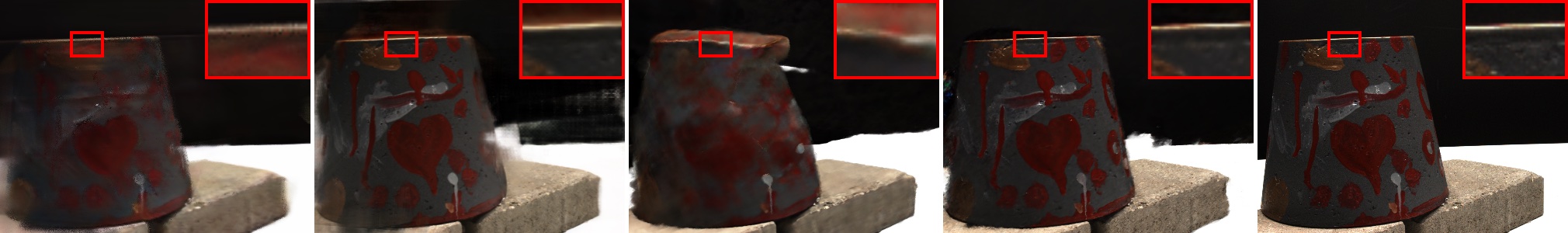}        %
         \caption{6 Input Views}
     \end{subfigure}
     \begin{subfigure}[b]{1.0\textwidth}
         \centering
        \includegraphics[width=\linewidth]{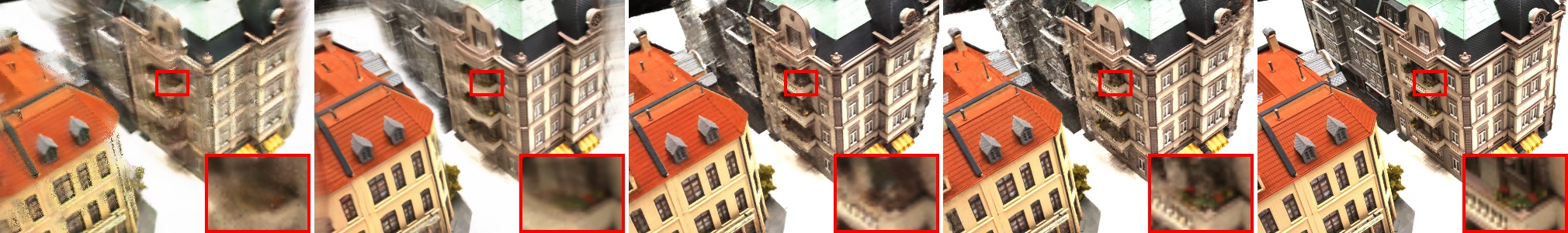}        
         \caption{9 Input Views}
     \end{subfigure}
     \vspace{-.7cm}
    \caption{
    \textbf{View Synthesis on DTU.}
    We show novel views generated by the baselines and our method for $3$, $6$ and $9$ input images. %
    While the baselines suffer from blurriness or incorrect scene geometry, our approach leads to sharp novel views.
    For $3$ input views, 
    DietNeRF leads to wrong geometry prediction and blends the input images rather than obtaining a 3D-consistent representation, due to the global nature of its semantic consistency loss.
    }
    \label{fig:dtu}
     \vspace{-.3cm}
\end{figure*}
\begin{figure*}
    \centering
    \begin{tabular}{P{0.1415\textwidth}P{0.1415\textwidth}P{0.1415\textwidth}P{0.1415\textwidth}P{0.1415\textwidth}P{0.1415\textwidth}}
     \scriptsize PixelNeRF~\cite{Yu2021CVPR} & \scriptsize PixelNeRF ft~\cite{Yu2021CVPR} & \scriptsize MVSNeRF ft~\cite{Chen2021ICCVa} & \scriptsize DietNeRF~\cite{Jain2021ICCV} & \scriptsize Ours & \scriptsize GT\\
    \end{tabular}
     \begin{subfigure}[b]{1.0\textwidth}
         \centering
        \includegraphics[width=\linewidth]{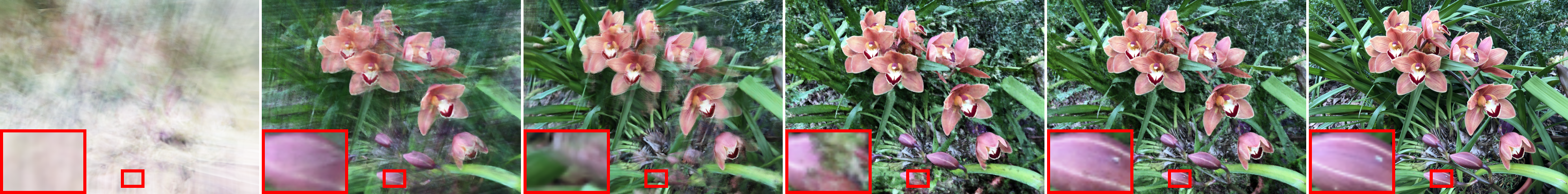}
         \caption{3 Input Views}
     \end{subfigure}
     \begin{subfigure}[b]{1.0\textwidth}
         \centering
        \includegraphics[width=\linewidth]{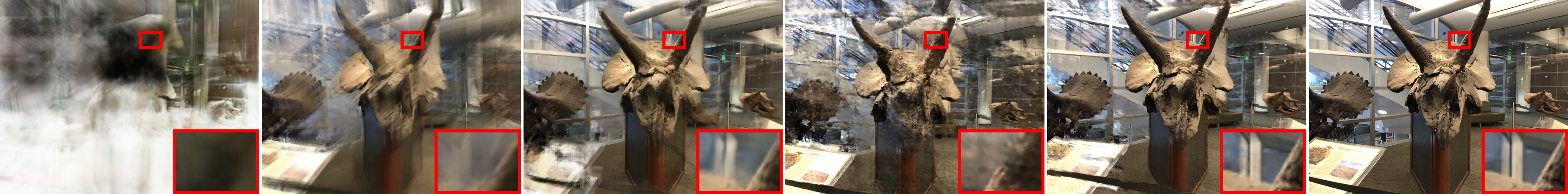}
         \caption{6 Input Views}
     \end{subfigure}
     \begin{subfigure}[b]{1.0\textwidth}
         \centering
        \includegraphics[width=\linewidth]{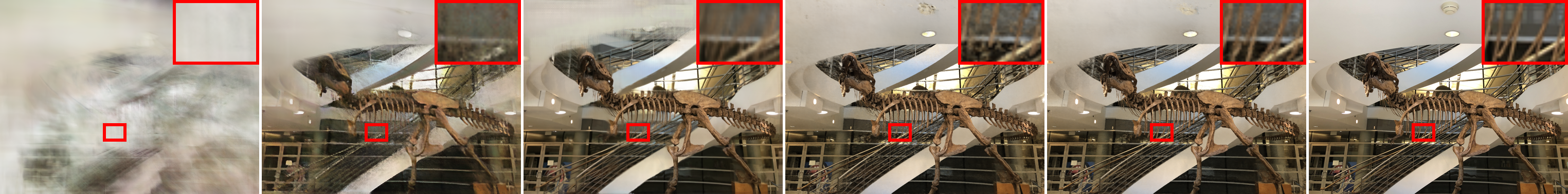}
         \caption{9 Input Views}
     \end{subfigure}
          \vspace{-.7cm}
    \caption{
    \textbf{View Synthesis on LLFF.}
    Conditional models overfit to the training data and hence perform poorly on test data from a novel domain. Further, novel views still appear slightly blurry despite additional fine-tuning (``ft''). 
    While DietNeRF does not require expensive pre-training similiar to our approach, our method leads to more accurate scene geometry, resulting in sharper and more realistic renderings.
    }
    \label{fig:llff}
         \vspace{-.6cm}
\end{figure*}

\boldpnov{DTU Dataset}
For $3$ input views, our method achieves quantitative results comparable to the best-performing conditional models (see~\tabref{tab:dtu}) which are pre-trained on other DTU scenes.
Compared to the other methods that also do not require pre-training, we achieve the best results. %
For $6$ and $9$ input views, our approach performs best compared to all baselines.
As evidenced by~\figref{fig:dtu}, we see that conditional models are able to predict good overall novel views, but become blurry particularly around edges and exhibit less consistent appearance for novel views whose cameras are far from the input views. For mip-NeRF and DietNeRF, which are not pre-trained (like our method), geometry prediction and hence synthesized novel views degrade for very sparse scenarios. Even with $6$ or $9$ input views, the results contain floating artifacts and incorrect geometry.
In contrast, our approach performs well across all scenarios, producing sharp results with more accurate scene geometry. %
  
\boldp{LLFF Dataset}
For conditional models, the LLFF dataset serves as an out-of-distribution scenario as the models are trained on DTU. We observe that SRF and PixelNeRF appear to overfit to the training data, which leads to low quantitative results (see~\tabref{tab:llff}). MVSNeRF generalizes better to novel data, and all three models benefit from additional fine-tuning.
For $3$ input views, mip-NeRF and DietNeRF are not able to generate competitive novel views. However, with $6$ or $9$ input views, they outperform the best conditional models.
Despite requiring fewer optimization steps than mip-NeRF and DietNeRF and no pre-training at all, our method achieves the best results across all scenarios.
From~\figref{fig:llff} we observe that the predictions from conditional models tend to be blurry for views far away from the inputs, and the test-time optimized baselines contain errors in predicted scene geometry.
Our method achieves superior geometry predictions and more realistic novel views.

\subsection{Ablation Studies}

\begin{table}
    \resizebox{\linewidth}{!}{\begin{tabular}{l|c|c|c|c}
\toprule
  &  PSNR $\uparrow$ & SSIM $\uparrow$ & LPIPS $\downarrow$ & Average $\downarrow$  \\
  \midrule
   w/o Scene Space Ann.\ & 10.17 & 0.613 & 0.332 & 0.291 \\
   w/o Geometry Reg.\ & \cellcolor{yellow!25}14.34 & \cellcolor{yellow!25}0.689 & \cellcolor{yellow!25}0.246 & \cellcolor{yellow!25}0.188 \\
   w/o Appearance Reg.\ & \cellcolor{orange!25}18.34 & \cellcolor{orange!25}0.742 & \cellcolor{orange!25}0.191 & \cellcolor{orange!25}0.117 \\
Ours & \cellcolor{red!25}18.89 & \cellcolor{red!25}0.745 & \cellcolor{red!25}0.190 & \cellcolor{red!25}0.112 \\
\bottomrule
\end{tabular}}
    \vspace{-.2cm}
    \caption{
    \textbf{Ablation Study.}
    We report our method ablating components on the DTU dataset for $3$ input views.
    For very sparse scenarios, we find that scene space annealing is crucial to avoid degenerate solutions. Further, regularizing scene geometry has a bigger impact on the performance than appearance regularization. Combining all components leads to the best performance.
    }
    \label{tab:ablation}
	\vspace{-.4cm}
\end{table}
In~\tabref{tab:ablation}, we ablate various components of our method.
For sparse inputs, we find that the proposed scene space annealing strategy avoids degenerate solutions. Further, regularizing geometry is more important than appearance, and combining all components leads to the best results.

\boldp{Ablation of Geometry Regularizer}
\begin{table}
    \resizebox{\linewidth}{!}{\begin{tabular}{l|c|c|c|c}
\toprule
  &  PSNR $\uparrow$ & SSIM $\uparrow$ & LPIPS $\downarrow$ & Average $\downarrow$  \\
  \midrule
Opacity Reg.\ \cite{Lombardi2019SIGGRAPH} & 11.07 & 0.617 & 0.309 & 0.268 \\
Ray Density Entropy Reg.\ & 13.93 & 0.680 & 0.254 & 0.198 \\
Normal Smooth.\ Reg.\ \cite{Niemeyer2020CVPR} & 14.22 & 0.683 & 0.251 & 0.193 \\
Density Surface Reg.\ & \cellcolor{yellow!25}14.71 & \cellcolor{yellow!25}0.687 & \cellcolor{yellow!25}0.247 & \cellcolor{yellow!25}0.184 \\Sparsity Reg.\ \cite{Hedman2021ICCV} & \cellcolor{orange!25}16.77 & \cellcolor{orange!25}0.711 & \cellcolor{orange!25}0.221 & \cellcolor{orange!25}0.145 \\Depth Smooth.\ Reg.\ (Ours) & \cellcolor{red!25}18.89 & \cellcolor{red!25}0.745 & \cellcolor{red!25}0.190 & \cellcolor{red!25}0.112 \\
\bottomrule
\end{tabular}}
    \vspace{-.2cm}
    \caption{
    \textbf{Geometry Regularization.}
    We compare different choices of geometry regularization strategies on DTU ($3$ input views) and find that our depth smoothness prior performs best.
    }
    \label{tab:ablation-geometry}
    \vspace{-.4cm}
\end{table}
In~\tabref{tab:ablation-geometry}, we investigate the performance of other geometry regularization techniques. %
We find that opacity-based regularizers (\eg, enforce rendered opacity values near to either $0$ or $1$) and density or normal smoothness priors (\eg minimize the distance between neighboring normal vectors in 3D), two strategies often used to enforce solid and smooth surfaces, do not produce accurate scene geometry.
Employing the sparsity prior from Hedman \etal~\cite{Hedman2021ICCV} leads to better quantitative results, but novel views still contain floating artifacts and the optimized geometry has holes. 
In contrast, our geometry regularization strategy
achieves the best performance. 
We hypothesize that similar to density-based~\cite{Mildenhall2020ECCV} vs.\ single surface optimization~\cite{Niemeyer2020CVPR, Sitzmann2019NIPS} for coordinate-based methods, providing gradient information along the full ray rather than a single point provides a more stable and informative learning signal.

\section{Conclusion}
\label{sec:discussion}
\begin{figure}
\captionsetup[subfigure]{justification=centering}
    \centering
     \begin{subfigure}[b]{1.0\linewidth}
         \centering
        \includegraphics[width=.32\linewidth]{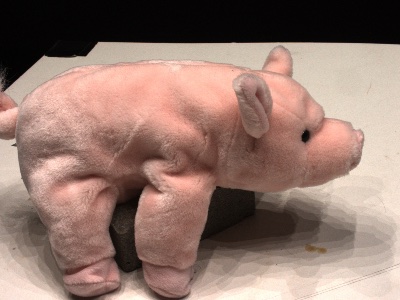}
        \includegraphics[width=.32\linewidth]{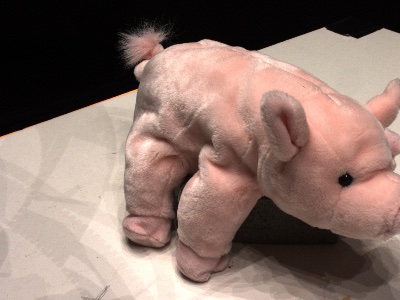}
        \includegraphics[width=.32\linewidth]{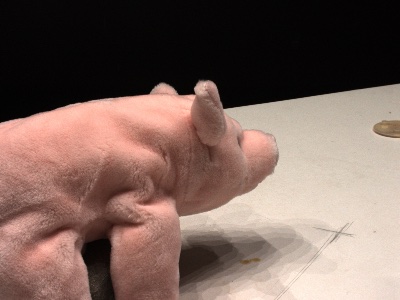}
        \caption{Sparse Set of 3 Input Views}\label{subfig:failure-input}
     \end{subfigure}
     \begin{subfigure}[t]{.49\linewidth}
         \centering
        \includegraphics[width=\linewidth]{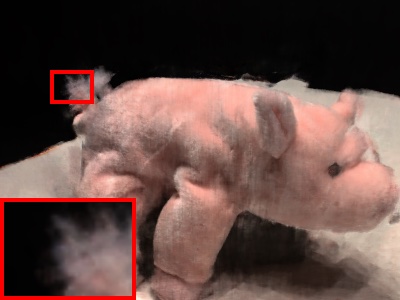}
        \caption{Prediction}\label{subfig:failure-ours}
     \end{subfigure}
     \begin{subfigure}[t]{.49\linewidth}
         \centering
        \includegraphics[width=\linewidth]{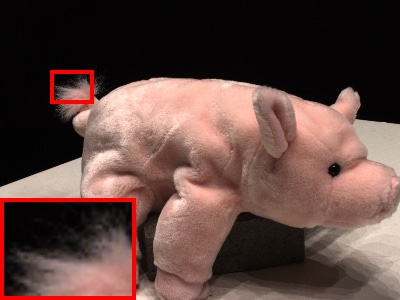}
    \caption{GT}\label{subfig:failure-gt}
     \end{subfigure}
     \vspace{-.1cm}
    \caption{
    \textbf{Failure Analysis.}
        As we do not attempt to hallucinate geometric details in this work, our model may lead to blurry predictions in unobserved regions with areas of fine geometry (\ref{subfig:failure-ours}).
        We identify incorporating uncertainty prediction or generative components into our model as interesting future work.
    }
    \label{fig:failure}
    \vspace{-.5cm}
\end{figure}

We have presented RegNeRF, a novel approach for optimizing Neural Radiance Fields (NeRF) in data-limited regimes.
Our key insight is that for sparse input scenarios, NeRF's performance drops significantly due to incorrectly optimized scene geometry and divergent behavior at the start of optimization. To overcome this limitation, we propose techniques to regularize the geometry and appearance of rendered patches from unseen viewpoints. In combination with a novel sample-space annealing strategy, our method is able to learn 3D-consistent representations from which high-quality novel views can be synthesized.
Our experimental evaluation shows that our model outperforms not only methods that, similar to us, only optimize over a single scene,  but in many cases also conditional models that are extensively pre-trained on large scale multi-view datasets.

\boldp{Limitations and Future Work}
In this work, we do not attempt to hallucinate geometric detail.
As a result, our model may lead to blurry predictions in unobserved areas with fine geometric structures (see~\figref{fig:failure}).
We identify incorporating uncertainty prediction mechanisms~\cite{Shen2021ARXIV} or generative components~\cite{Schwarz2020NEURIPS,Niemeyer2021CVPR,Chan2021CVPR, Gu2021ARXIV} as promising future work.

\FloatBarrier

{\small
\bibliographystyle{ieee_fullname}
\bibliography{bibliography_long,bibliography,bibliography_custom}
}

\end{document}